\documentclass[conference]{IEEEtran}
\usepackage{cite}
\usepackage{amsmath,amssymb,amsfonts}
\usepackage{algorithmic}
\usepackage{graphicx}
\usepackage{textcomp}
\usepackage[colorlinks=true, linkcolor=black, citecolor=blue, urlcolor=blue]{hyperref}
\usepackage{siunitx}
\usepackage{balance}
\usepackage{fancyhdr}

\makeatletter
\let\MYcaption\@makecaption
\makeatother

\usepackage[font=footnotesize]{subcaption}

\makeatletter
\let\@makecaption\MYcaption
\makeatother

\def\BibTeX{{\rm B\kern-.05em{\sc i\kern-.025em b}\kern-.08em
    T\kern-.1667em\lower.7ex\hbox{E}\kern-.125emX}}
\begin{document}

\title{\huge TEX-CUP:\\The University of Texas Challenge for Urban Positioning}

\makeatletter
\newcommand{\linebreakand}{%
  \end{@IEEEauthorhalign}
  \hfill\mbox{}\par
  \mbox{}\hfill\begin{@IEEEauthorhalign}
}
\makeatother

\author{
\IEEEauthorblockN{Lakshay Narula, Daniel M. LaChapelle, Matthew J. Murrian, J. Michael Wooten, Todd E. Humphreys}
\IEEEauthorblockA{\textit{Radionavigation Laboratory} \\
\textit{The University of Texas at Austin}\\
Austin, TX, USA}
\vspace{1pt}
\linebreakand
\IEEEauthorblockN{Elliot de Toldi, Guirec Morvant, Jean-Baptiste Lacambre}
\IEEEauthorblockA{\textit{iXblue INC} \\
Denver, CO, USA}
}

\maketitle

\begin{abstract}

  A public benchmark dataset collected in the dense urban center of the city of
  Austin, TX is introduced for evaluation of multi-sensor GNSS-based urban
  positioning.  Existing public datasets on localization and/or odometry
  evaluation are based on sensors such as lidar, cameras, and radar. The role
  of GNSS in these datasets is typically limited to the generation of a
  reference trajectory in conjunction with a high-end inertial navigation
  system (INS). In contrast, the dataset introduced in this paper provides raw
  ADC output of wideband intermediate frequency (IF) GNSS data along with
  tightly synchronized raw measurements from inertial measurement units (IMUs)
  and a stereoscopic camera unit. This dataset will enable optimization of the
  full GNSS stack from signal tracking to state estimation, as well as sensor
  fusion with other automotive sensors. The dataset is available at
  \url{http://radionavlab.ae.utexas.edu} under Public Datasets. Efforts to
  collect and share similar datasets from a number of dense urban centers
  around the world are under way.

\end{abstract}


\newif\ifpreprint
\preprintfalse
\preprinttrue

\ifpreprint

\pagestyle{plain}
\thispagestyle{fancy}  
\fancyhf{} 
\renewcommand{\headrulewidth}{0pt}
\rfoot{\footnotesize \bf May 2020 preprint of paper accepted for publication} \lfoot{\footnotesize \bf
  Copyright \copyright~2020 by Lakshay Narula}

\else

\thispagestyle{empty}
\pagestyle{empty}

\fi

\section{Introduction}

\IEEEPARstart{D}{evelopment} of automated ground vehicles (AGVs) has spurred
research in lane-keeping assist systems, automated intersection
management~\cite{fajardo2011automated}, tight-formation platooning, and
cooperative sensing~\cite{choi2016mmWaveVehicular, lachapelle2020riskIcassp},
all of which demand accurate (e.g., 50-cm at 95\%) ground vehicle positioning
in an urban environment.  But the majority of positioning techniques, and the
associated performance benchmarks, developed thus far are based on lidar or
cameras, which perform poorly in low-visibility conditions such as snowy
whiteout, dense fog, or heavy rain. Adoption of AGVs in many parts of the world
will require all-weather localization techniques.

Radio-wave-based sensing techniques such as radar and GNSS remain operable even
in extreme weather conditions~\cite{yen2015evaluation} because their
longer-wavelength electromagnetic radiation penetrates snow, fog, and rain.
Carrier-phase-differential GNSS (CDGNSS), also known as real time kinematic
(RTK) GNSS, has been successfully applied for the past two decades as an
all-weather decimeter-accurate localization technique in open-sky conditions.
Similarly, inertial sensing techniques are also unaffected by weather
conditions. A combination of low-cost inertial- and radio-based localization is
a promising direction towards precise all-weather urban positioning for AGVs.

While application of CDGNSS/RTK techniques for urban positioning has previously
been limited due to expensive coupling with tactical grade
IMUs~\cite{petovello2004benefits}, recent work has shown that 20-cm-accurate
(95\%) low-cost unaided CDGNSS positioning is possible at 87\% availability
with dual-frequency GPS and Galileo signals, even in the dense urban downtown
of Austin, TX~\cite{humphreys2019deepUrbanIts}.  Similarly,~\cite{li2018high}
shows that unaided dual-frequency GPS-, BeiDou-, and GLONASS-based CDGNSS
positioning can achieve decimeter-accurate (95\%) positioning rate of 76.7\% on
a 1-hour drive along an urban route in Wuhan, China, and the availability can
be further improved to 86.1\% after integration with a MEMS IMU. Meanwhile,
recent urban CDGNSS evaluation of commercial receivers
in~\cite{jackson2018assessmentRtk} indicates that no low-to-mid-range consumer
CDGNSS solution offers greater than 35\% decimeter-accurate solution
availability in urban areas, despite a dense reference network and
dual-frequency capability.

Similarly, until recently precise point positioning (PPP) algorithms required a
long convergence time and as such were limited to surveying applications. With
the proliferation of the number of GNSS satellites and better numerical models
for atmospheric corrections~\cite{banville2014global}, recent efforts have
reported instantaneous convergence times for PPP in open sky or light urban
conditions.  The authors predict that efforts towards accuracy and availability
of PPP in urban areas will be soon forthcoming.

The concern with development of GNSS-based precise positioning techniques as
described above is that different algorithms may have been evaluated on
datasets of different difficulties, even if the general environment may be
described as urban. As an extreme example, consider the data collection route
presented in Fig.~\ref{fig:route}. Over the entirety of the dataset, a
Septentrio AsteRx4 RTK receiver used as a part of this data collection reports
an integer-ambiguity-fixed RTK solution at 70.6\% of all epochs. However,
as is typical, the data collection routine involved $\approx$\SI{10}{\minute}
stationary open-sky periods at the beginning and end of data collection.
Excluding these periods brings down the fixed solution availability to 49.6\%
on the remainder of the dataset. In fact, when restricted strictly to the dense
urban southern portion of the test route, the availability of reported precise
RTK solutions is only 21.3\%. As such, it is at best challenging, and at worst
misleading, to compare precise GNSS positioning algorithms on different
datasets. Additionally, multipath properties of the GNSS antenna and phase
stability of the sampling clock are other important factors that likely affect
the performance analysis.  In the opinion of the authors, the precise GNSS
positioning community must converge on a shared and challenging dataset to
evaluate their algorithms and thereby identify the critical components of a
robust and accurate urban positioning engine.

As a precedent, similar benchmarks such as the KITTI
dataset~\cite{Geiger2013IJRR} for visual odometry and object segmentation, and
the ImageNet dataset~\cite{deng2009imagenet} for object instance recognition
have served greatly towards the progress of their respective communities. The
Oxford Robotcar Dataset~\cite{RobotCarDatasetIJRR} has been a similarly
important benchmark dataset in the field of repeatable ground vehicle
localization with lidars and cameras. However, none of the existing robotic
localization datasets are focused on GNSS-based precise urban localization. The
dataset being introduced in this paper addresses this gap for the GNSS research
community.

The goal of the University of Texas Challenge for Urban Positioning is twofold:
to enable the precise GNSS positioning community to evaluate and compare a
variety of existing and upcoming techniques on a shared and challenging
benchmark, and to save the time and effort required to assemble a high-quality
data recording platform for urban positioning research.

\section{Sensor Platform}

The roving dataset is captured with an integrated perception platform named
the University of Texas \emph{Sensorium}, shown in Figs.~\ref{fig:sensorium-external}
and~\ref{fig:sensorium-internal}, equipped with the following sensors:
\begin{itemize}
  \item \num{2}$\times$ Antcom G8Ant-3A4TNB1 high performance GNSS patch
    antennas (NGS code: ACCG8ANT\_3A4TB1). Triple frequency L1/L2/L5;
    \SI{40}{\decibel} low-noise amplifier.
  \item \num{1}$\times$ RadioLynx GNSS RF front end. Dual frequency L1/L2;
    \SI{5}{Msps} bandwidth on both channels; support for two GNSS antennas;
    developed in-house; provided with Bliley LP-62 low-power
    \SI{10}{\mega\hertz} OCXO external reference.
  \item \num{1}$\times$ NTLab B1065U1-12-X configurable RF front end.
    Configured to capture L1/L2/L5 signals from one GNSS antenna with a wide
    bandwidth of \SI{53}{Msps}; provided with Bliley LP-62 low-power
    \SI{10}{\mega\hertz} OCXO external reference.
  \item \num{1}$\times$ u-blox EVK-M8T. Single-frequency (L1)
    multi-constellation mass-market receiver.
  \item \num{1}$\times$ Bosch BMX055 9-axis IMU. Low-cost MEMS device;
    smartphone-grade IMU noise characteristics; \SI{150}{\hertz} output rate.
  \item \num{1}$\times$ LORD MicroStrain 3DM-GX5-25 AHRS. High-performance MEMS
    device; industrial-grade IMU noise characteristics; \SI{100}{\hertz} output
    rate.
  \item \num{2}$\times$ Basler acA2040-35gm cameras. $2048 \times 1536$
    resolution; monochromatic; Sony IMX265 CMOS sensor; global shutter;
    hardware triggered at \SI{10}{fps}; $\approx$\SI{50}{\centi\meter}
    baseline; Kowa LMVZ4411 lenses.
  \item \num{1}$\times$ Delphi ESR 2.5 (24VDC) L2C0051TR electronically
    scanning radar.  Simultaneous mid- and long-range measurement modes;
    mid-range \SI{60}{\meter}, \SI{90}{\degree} field-of-view; long-range
    \SI{174}{\meter}, \SI{20}{\degree} field-of-view; \SI{20}{\hertz} scan
    rate.
  \item \num{2}$\times$ Delphi SRR2 single beam monopulse radars. Range
    \SI{80}{\meter}; field-of-view \SI{150}{\degree}; \SI{20}{\hertz} scan
    rate. When mounted as shown in Figs.~\ref{fig:sensorium-external}
    and~\ref{fig:sensorium-internal}, the three radars provide
    \SI{210}{\degree} of coverage around the vehicle.
  \item \num{1}$\times$ Taoglas 4G LTE MIMO antenna. Provides connectivity to
    the network for CDGNSS corrections.
\end{itemize}
For the purposes of this evaluation dataset, the Sensorium is equipped with an
iXblue ATLANS-C: a high-performance RTK-GNSS coupled fiber-optic gyroscope
(FOG) INS (not shown in Figs.~\ref{fig:sensorium-external}
and~\ref{fig:sensorium-internal}). The Septentrio AsteRx4 RTK receiver inside
the ATLANS-C is attached to one of the two GNSS antennas, and tracks most
constellations on all three GNSS frequencies. The post-processed
fused RTK-INS position solution obtained from the ATLANS-C is taken to be the
ground truth trajectory.

\begin{figure}[ht]
  \centering
  \includegraphics[width=\linewidth,trim={50 100 50 125},clip]{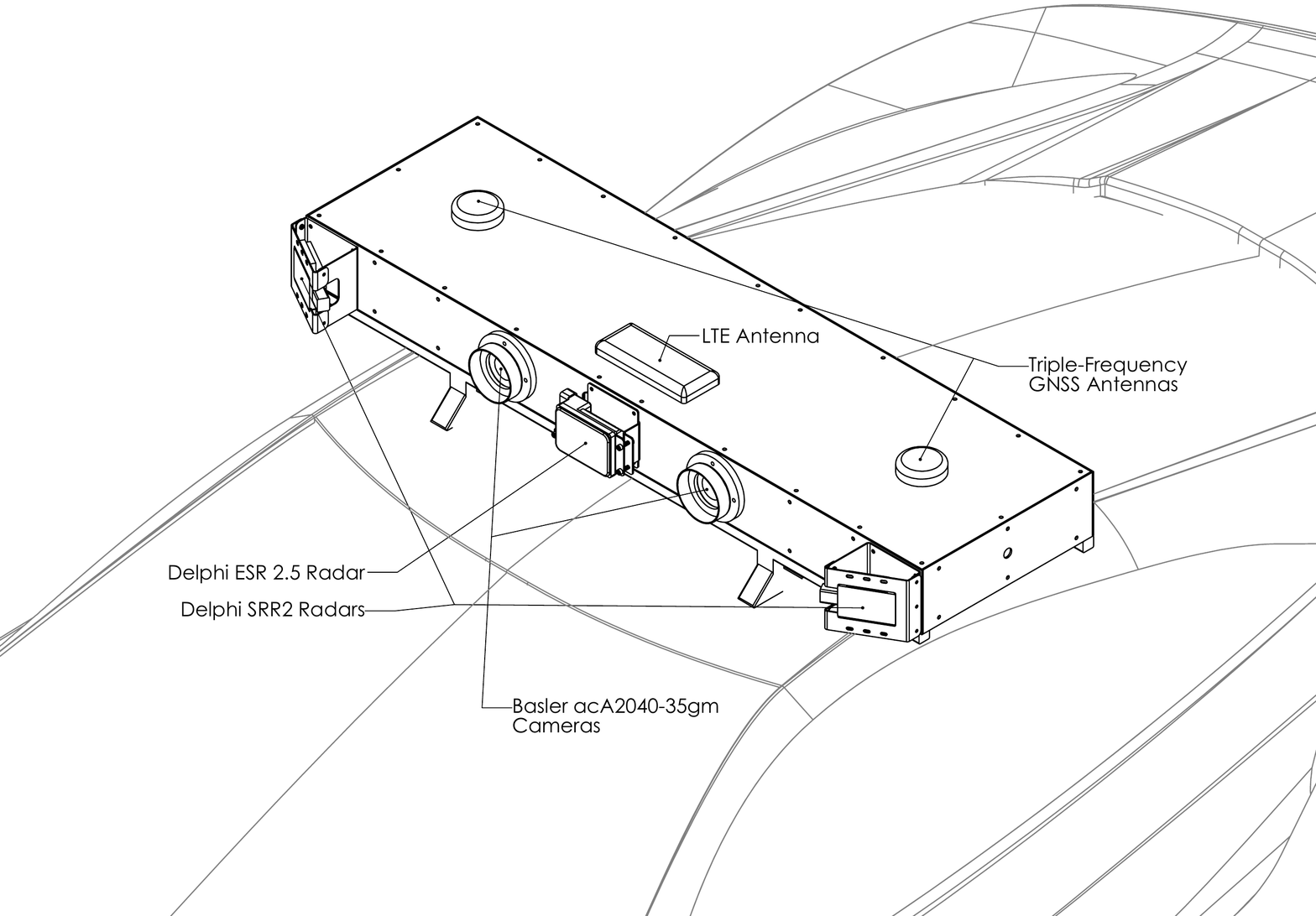}
  \caption{The University of Texas Sensorium is a platform for automated and
  connected vehicle perception research. The Sensorium features two L1/L2/L5
  GNSS antennas, wideband GNSS RF front ends, smartphone- and industrial-grade
  MEMS IMUs, stereoscopic cameras, automotive radars, and LTE connectivity.}
  \label{fig:sensorium-external}
\end{figure}

\begin{figure}[ht]
  \centering
  \includegraphics[width=\linewidth,trim={100 75 75 150},clip] {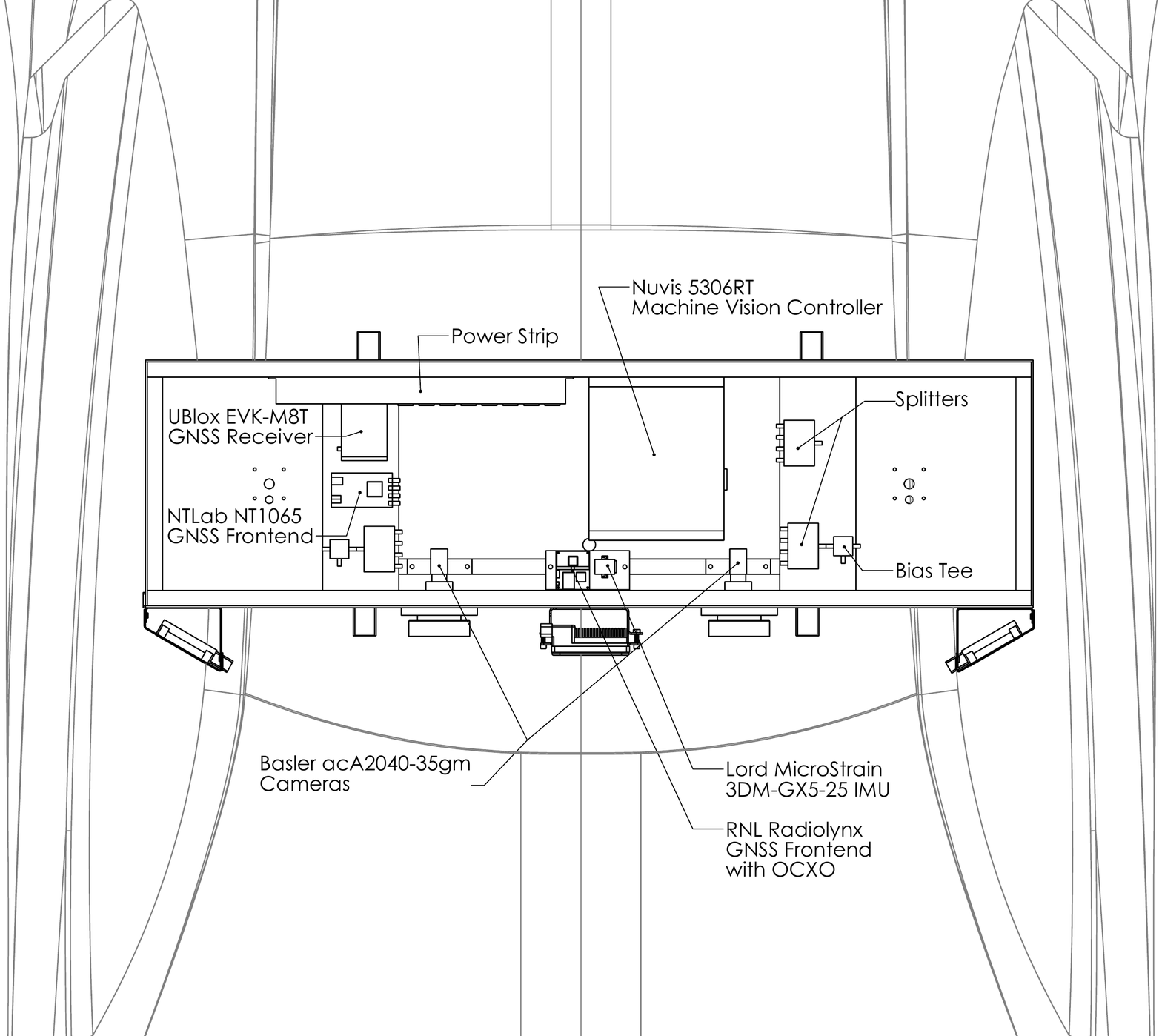}
  \caption{Inside view of the University of Texas Sensorium, showing the
  internal organization of a desktop-class computer, IMUs, two GNSS RF front
  ends, and a stereoscopic camera setup.}
  \label{fig:sensorium-internal}
\end{figure}

The Sensorium houses a rugged Nuvis N5306RT computer with a modest
desktop-level configuration. The computer runs Ubuntu Linux and logs data from
all sensors and devices. Most data logging processes are developed in-house for
precise synchronization between sensor data. Details on sensor synchronization
are provided in Sec.~\ref{sec:sync}.

To enable CDGNSS-based positioning, the dataset also includes GNSS data logged
from a nearby reference antenna with a clear view of the sky. The reference
antenna is a geodetic-grade Trimble Zephyr II (NGS code: TRM57971.00). For
consistency with the rover, raw IF reference data is logged with identical
RadioLynx and NTLab RF front ends. For completeness, RINEX-format reference
data from an identical Septentrio AsteRx4 receiver is also logged. The rover
platform is always within \SI{4}{\kilo\meter} of the reference antenna,
representing ideal CDGNSS conditions.

\section{Data Collection}

The test route, depicted in Fig.~\ref{fig:route}, runs the gamut of
light-to-dense urban conditions, from open-sky to narrow streets with
overhanging trees to the high-rise urban city center.

The data capture begins and ends with a stationary interval of several minutes
in open sky conditions to allow confident bookending for the ground truth
system. The first part of the trajectory runs through the semi-urban conditions
north of the University of Texas campus, passing under two pedestrian bridges.
The second part of the trajectory passes through an area with narrow streets
lined by tall residential apartment buildings and dense foliage. The rest of
the test route combs through the dense urban center of the city of Austin, TX,
driving through every east-west street in the city downtown.

\begin{figure}[ht]
  \centering
  \includegraphics[width=\linewidth,trim={300 0 300 0},clip] {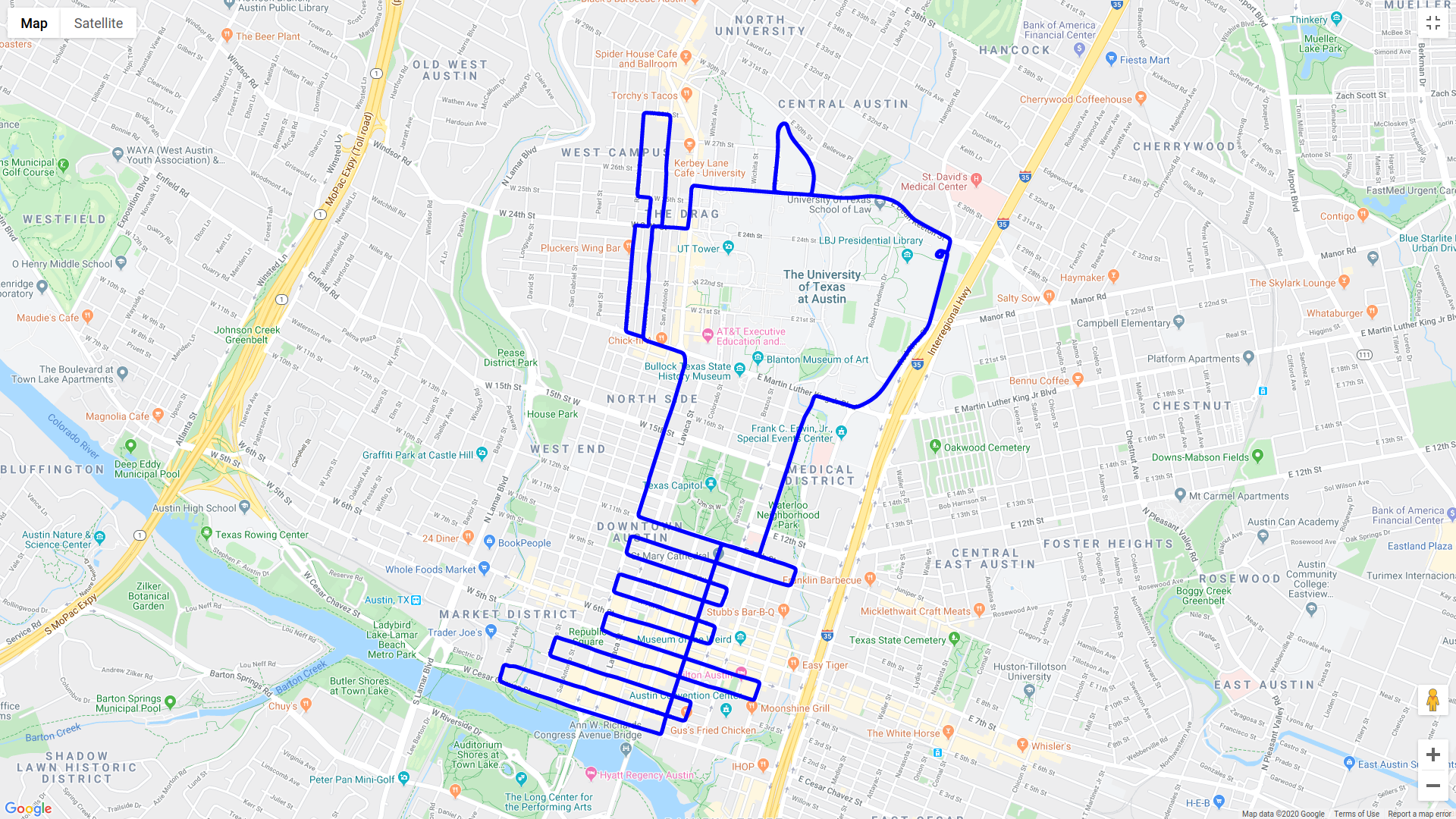}
  \caption{Test route through The University of Texas west campus and Austin
  downtown. These areas are the most challenging for precise GNSS-based
  positioning. The route was driven once on a weekday and again on the weekend
  to evaluate robustness of mapping-based methods to changes in traffic and
  parking patterns.}
  \label{fig:route}
\end{figure}

The number of signals tracked by a receiver is a good indicator of the level of
difficulty posed by the dataset. Fig.~\ref{fig:nsignals} shows two extremes of
this metric by comparing the low-cost mass-market u-blox M8T and the
high-performance all-in-view Septentrio AsteRx4. As mentioned before, the u-blox
M8T is a single frequency receiver, and is only able to track GPS and GLONASS
signals in the presented dataset. The number of tracked signals during the
\SI{30}{\minute} challenging downtown portion of the dataset is under \num{15}
for the M8T, making it unlikely to produce reliable CDGNSS position
estimates~\cite{humphreys2019deepUrbanIts}.  At the other end of the
performance spectrum, the AsteRx4 receiver is a state-of-the-art all-in-view
receiver, tracking all constellations in all GNSS bands. For this receiver, the
number of tracked signals is above \num{20} for most of the challenging portion
of the dataset.

\begin{figure}[ht]
  \centering
  \includegraphics[width=\linewidth] {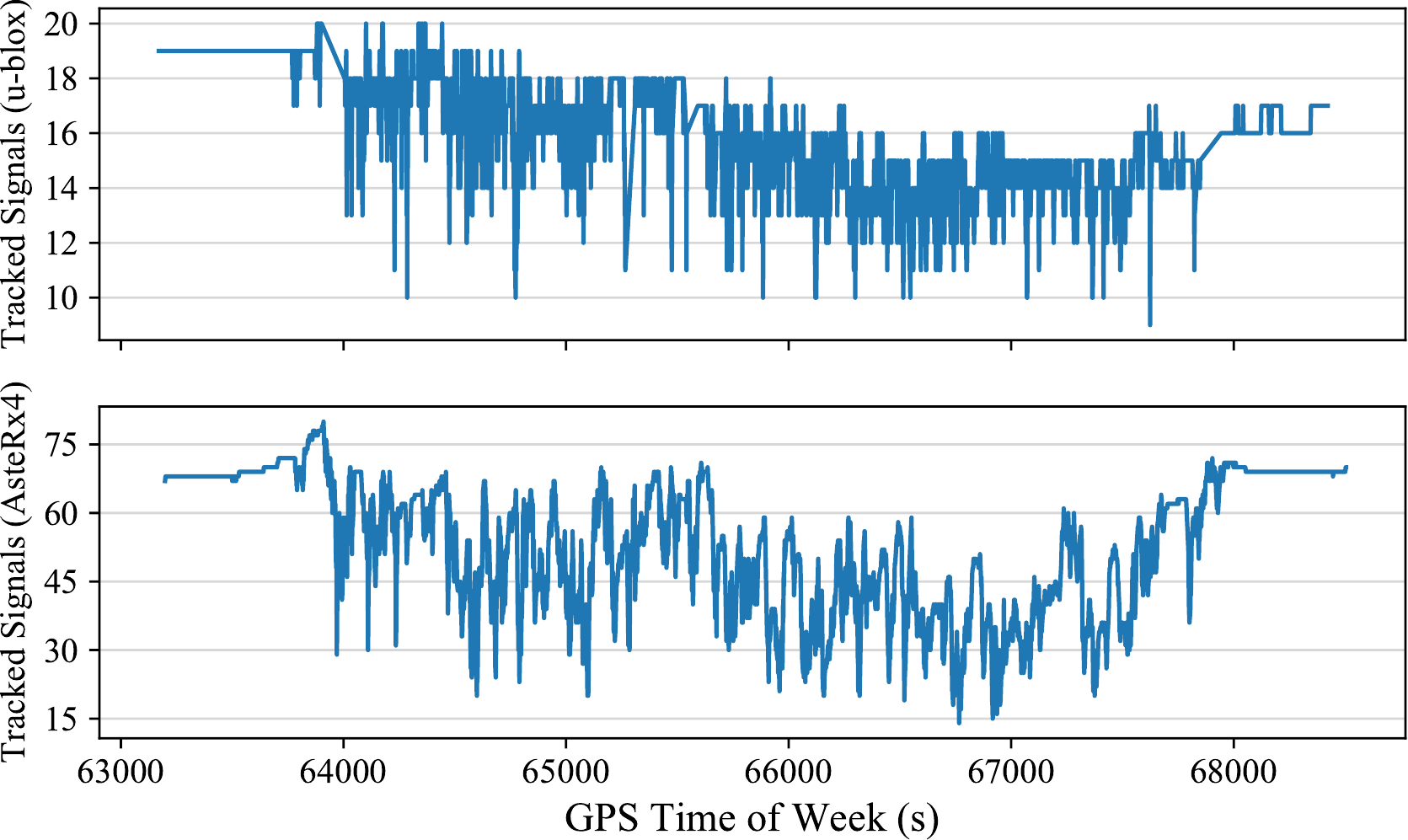}
  \caption{A comparison of the number of tracked signals over the duration of
  the dataset for the u-blox M8T receiver (top) and the Septentrio AsteRx4
  receiver (bottom). The u-blox M8T is an L1-only receiver tracking GPS and
  GLONASS signals. The Septentrio AsteRx4 is a triple-frequency all-in-view
  receiver.}
  \label{fig:nsignals}
\end{figure}

Fig.~\ref{fig:atx-dataset} shows Google Street View imagery from the driven
route for a qualitative assessment of the dataset difficulty. A KML file with
the full route is provided along with the dataset for easy visualization of the
urban conditions.

\begin{figure*}[htb!]
  \centering
  \begin{minipage}[b]{0.325\textwidth}
    \centering
    \includegraphics[width=\linewidth,draft=false]{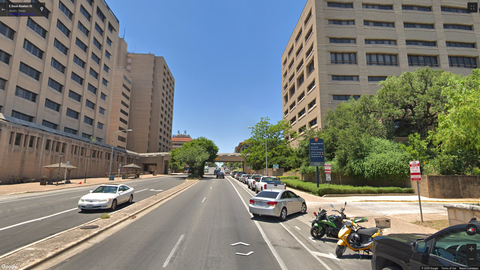}
  \end{minipage}
  \begin{minipage}[b]{0.325\textwidth}
    \centering
    \includegraphics[width=\linewidth,draft=false]{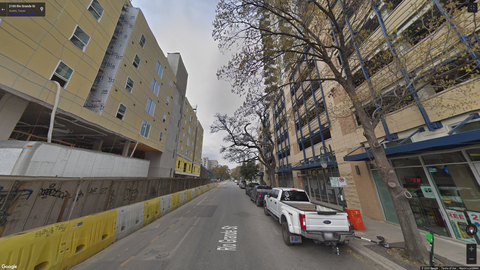}
  \end{minipage}
  \begin{minipage}[b]{0.325\textwidth}
    \centering
    \includegraphics[width=\linewidth,draft=false]{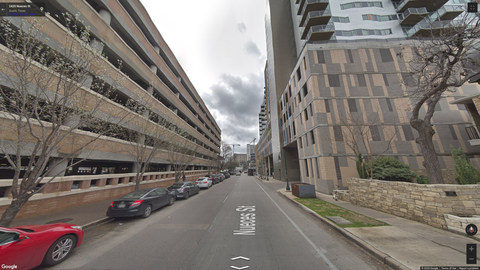}
  \end{minipage} \par\medskip
  \begin{minipage}[b]{0.325\textwidth}
    \centering
    \includegraphics[width=\linewidth,draft=false]{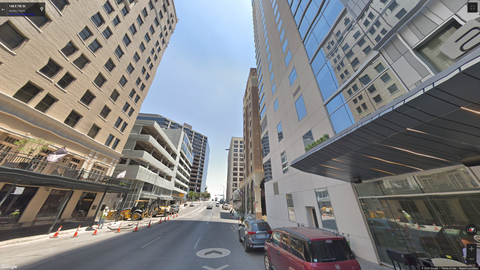}
  \end{minipage}
  \begin{minipage}[b]{0.325\textwidth}
    \centering
    \includegraphics[width=\linewidth,draft=false]{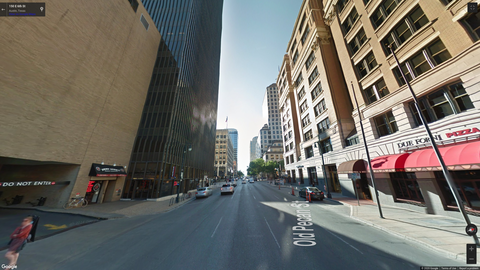}
  \end{minipage}
  \begin{minipage}[b]{0.325\textwidth}
    \centering
    \includegraphics[width=\linewidth,draft=false]{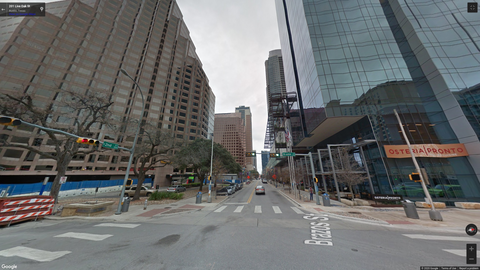}
  \end{minipage} \par\medskip
  \begin{minipage}[b]{0.325\textwidth}
    \centering
    \includegraphics[width=\linewidth,draft=false]{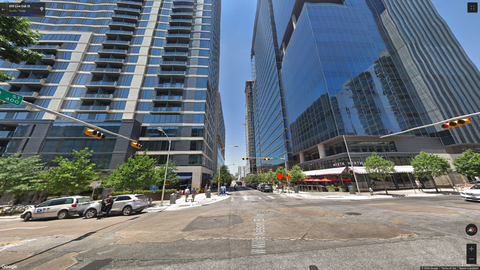}
  \end{minipage}
  \begin{minipage}[b]{0.325\textwidth}
    \centering
    \includegraphics[width=\linewidth,draft=false]{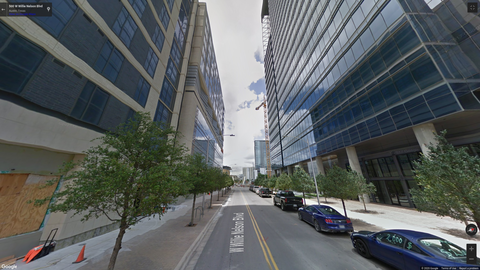}
  \end{minipage}
  \begin{minipage}[b]{0.325\textwidth}
    \centering
    \includegraphics[width=\linewidth,draft=false]{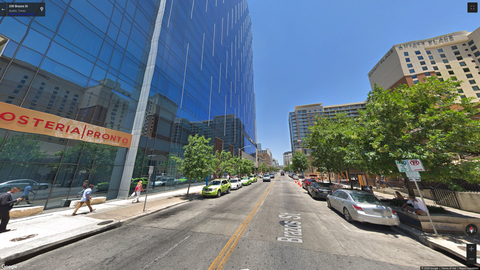}
  \end{minipage}
  \caption{Google Street View imagery of a few challenging scenarios
  encountered in the dataset.}
  \label{fig:atx-dataset}
\end{figure*}

The trajectory shown in Fig.~\ref{fig:route} is driven twice, once on
Thursday, May 9, 2019, and again on Sunday, May 12, 2019. The repeated
trajectory enables use of mapping-based techniques and their evaluation with
sufficient variation in traffic and parking patterns between a weekday and a
weekend.

\subsection{Data Formats}

This section describes the formats of different sensor data made available as
part of this dataset. The description is organized by the different devices
generating the data.

\subsubsection{RadioLynx Front End}

The RadioLynx RF front end generates two-bit-quantized samples from two
antennas at the rover and a single antenna at the reference station, capturing
\SI{4.2}{\mega\hertz} bandwidth at both L1 and L2 bands around the GPS
frequencies. The raw IF data from the three antennas is made available in a
binary format documented along with the dataset, including the required IF
parameters. Raw IF data enable development of new signal tracking strategies
for urban precise positioning, and allow high-sensitivity receivers to track
weak signals that may not have been tracked by the receivers in the recording
platform. Raw IF samples from the reference antenna can be used for data bit
wipeoff~\cite[Sec.  III-D]{humphreys2019deepUrbanIts}, if desired.

The dataset also provides tracked pseudorange and carrier-phase observables
generated by the GRID software-defined
receiver~\cite{humphreys2019deepUrbanIts} operating on the RadioLynx raw IF
samples for both rover antennas and the reference antenna. At the time of
writing, the GRID receiver tracks GPS, Galileo, and SBAS signals. The
observables are provided in the RINEX format.

\subsubsection{NTLab Front End}

The NTLab RF front end produces two-bit-quantized samples from one of the two
rover antennas and the antenna at the reference station. With a sample rate of
\SI{79.5}{\mega\hertz}, the NTLab front end captures signals at L1, L2, and L5
frequencies with a wide bandwidth. The raw IF data from both the rover and
reference antennas are made available at this time, tracked observables in
RINEX format will be made available soon.

\subsubsection{Septentrio AsteRx4}

The Septentrio AsteRx4 receiver housed inside iXblue ATLANS-C produces
observables for GPS, Galileo, GLONASS, BeiDou, and SBAS at all three GNSS
frequencies. All these observables are made available in RINEX format.

\subsubsection{u-blox M8T}

The NMEA output from the u-blox M8T receiver is provided with the dataset for
comparison to a competitive mass-market receiver.

\subsubsection{Stereo Cameras}

Timestamped stereo images from the two Basler cameras are made available in
HDF5 format. As detailed later in Sec.~\ref{sec:sync}, camera images are
timestamped by the Sensorium computer when the image is received over Ethernet.
The dataset also provides the exposure time for individual images if it may be
desirable to account for its variation.

Accurate intrinsic and extrinsic calibration of cameras is important for
camera-based positioning. This dataset provides an HDF5 archive of stereo and
monocular calibration images captured with the Sensorium before the data
capture, along with measurements of the calibration patterns. These archives
may be used to obtain both intrinsic and extrinsic calibration parameters as
required, e.g., using the Kalibr calibration toolbox~\cite{furgale2013unified}.

\subsubsection{Bosch IMU}

To evaluate the benefit of low-cost inertial aiding in urban areas, the dataset
includes timestamped specific force, angular rate, and temperature measurements
from the Bosch BMX055 IMU in CSV format. This IMU is built-in to the RadioLynx
board, and has been set up such that the IMU data timestamps can be traced back
to the GNSS RF sampling clock. This enables highly accurate correspondence
between the IMU timestamps and GPS time.

\subsubsection{LORD MicroStrain IMU}

Timestamped specific force and angular rate measurements from the
high-performance LORD MicroStrain MEMS IMU are made available in CSV format.
The LORD IMU accepts a PPS (pulse per second) signal generated by the u-blox
receiver to synchronize to GPS time. LORD IMU measurements are internally
compensated for temperature variation.

\subsubsection{ATLANS-C IMU}

The dataset includes specific force and angular rate measurements from the
highly stable accelerometers and FOGs housed in the iXblue ATLANS-C. These data
are only made available from the May 9, 2019 data collection session. The
ATLANS-C data from May 12, 2019 data are held back for performance evaluation.

\subsubsection{Ground Truth Trajectory}

A trustworthy ground truth trajectory against which to compare the reported
trajectory of a system under test is indispensable for urban positioning
evaluation. Post-processing software provided by iXblue generates a
forward-backward smoothed position and orientation solution with fusion of
AsteRx4 RTK solutions and inertial measurements. The post-processed solution is
accurate to better than \SI{20}{\centi\meter} throughout the dataset, and may
be considered as the ground truth trajectory. As with the ATLANS-C IMU
measurements, the ground truth trajectory is only made available from the May 9,
2019 session. The ground truth trajectory from May 12, 2019 is withheld for
evaluation of community solutions. The authors may advertise the performance of
community submissions on the dataset webpage with consent from the developer.

\subsection{Interface with Receivers}

The dataset is easiest to interface to with a software-defined receiver, since
these receivers typically accept a stream of digitized IF samples as the input.
For receivers that only accept RF input, it may be possible to replay the
provided raw IF samples after upconversion to RF with use of a GNSS
replay/playback system similar to LabSat 3 Wideband~\cite{labsat3wideband}.

\subsection{Planned Worldwide Extension}

In partnership with iXblue, TEX-CUP is currently being extended to include raw
GNSS IF and IMU data from various worldwide dense urban centers. These future
data captures will use a simplified version of the Sensorium rover platform,
including the same NTLab and RadioLynx front ends, a newer u-blox receiver
(ZED-F9P), as well as the Septentrio AsteRx4 and ATLANS A7 (upgraded version of
the ATLANS-C) or ATLANS A9 (best in class) INS for the ground truth trajectory.
Raw IMU data from the Bosch BMX055 and ATLANS will also be included.

Urban centers currently under consideration for future data collection include
Denver, CO, Boston, MA, and San Diego, CA in the US, and Paris, Amsterdam,
Singapore, and Beijing internationally.

\section{Sensor Calibration \& Synchronization}
Accurate calibration and synchronization of all sensors is critical for any
localization dataset. The performance of GNSS/INS, odometry, and SLAM
techniques strongly depends on the accuracy of sensor calibration and
synchronization.

\subsection{Intrinsic Calibration}
\label{sec:intrinsic-calibration}

Intrinsic sensor calibration is necessary for cameras and IMUs, while antenna
and front-end calibration may be beneficial in high-accuracy and
high-availability GNSS applications.

\subsubsection{Cameras}
Intrinsic camera calibration may be performed by capturing images of a known
calibration pattern at different scales and orientations. The dataset includes
such a capture for the Sensorium cameras. These images may be used with a tool
such as Kalibr~\cite{furgale2013unified} to obtain intrinsic camera parameters
including focal length, principal point, lens distortion, etc. It must be noted
that platform vibrations during data collection can lead to small variations in
the intrinsic calibration parameters. It is most desirable to continuously
track the calibration parameters in real time in combination with CDGNSS and/or
IMUs.

\subsubsection{IMUs}
The dataset provides a \SI{24}{\hour} long stationary capture of IMU
measurements from the Bosch and LORD IMUs to enable calibration of IMU noise
and bias stability parameters.  In addition to noise and bias stability, IMU
intrinsic calibration involves estimation of accelerometer and gyroscope biases
and scale factors.  Unfortunately, \emph{a priori} intrinsic calibration is
typically not feasible due to variable turn-on-to-turn-on bias properties of
the IMUs.  It is thus common to track the IMU bias and scale factor parameters
in combination with GNSS and/or vision-based
positioning~\cite{kok2017inertial}.

\subsubsection{GNSS Antennas}
Intrinsic calibration of the Sensorium's two GNSS antennas amounts to
developing a model for antenna phase center variations (PCVs) as a function of
the direction of arrival of an incoming signal.  Such a calibration can be
obtained at the carrier phase level either relative to a reference antenna, as
in \cite{mader1999gps}, or in absolute terms, as in \cite{bilich2010gnss}.  The
U.S. National Geodetic Survey (NGS) offers absolute calibration files for a
wide variety of antenna models, including for the type of antenna on the
Sensorium (NGS code: ACCG8ANT\_3A4TB1) and at the reference station (NGS code:
TRM57971.00)\footnote{See \url{https://www.ngs.noaa.gov/ANTCAL/index.xhtml}}.
Users of the TEX-CUP data will tend to see improved CDGNSS availability and
accuracy when these calibrations are applied.

PCV models such as offered by the NGS cannot, however, compensate for local
effects: The Sensorium's antennas are mounted on a broad aluminum backplane
that affects the antennas' PCV behavior in a way not captured by the NGS model.
To obtain a more accurate PCV model for use with the TEX-CUP data set, a
relative calibration was performed between each of the Sensorium's antennas
\emph{in situ} and the TEX-CUP reference antenna.  GNSS phase and pseudorange
observables were collected over a two-day period and a PCV modeling procedure
like the one presented in \cite{mader1999gps} was performed, except that the
model was based on double- rather than single-difference carrier phase
measurements.  Let $z$ represent the zenith angle (angular departure from the
antenna boresight), and $a$ represent the azimuth angle of an incoming signal,
both in radians.  Then the additional carrier phase length, in meters, for a
signal arriving from direction $(z,a)$ is modeled as
\begin{align*}
  d(z, a) = \left(\sum_{i = 1}^n g_i z^i\right)\left[1 +
  \sum_{j = 1}^m g_{cj}\cos(ja) + g_{sj}\sin(ja)\right]
\end{align*}
Azimuthal coefficients $g_{cj}$ and $g_{sj}$, and elevation coefficients $g_i$,
are obtained via a nonlinear least squares fitting procedure.  Separate sets of
coefficients may be obtained for each of the antennas' three receiving
frequencies.

The azimuthal coefficients were found to be too small to be estimated reliably
from the two-day data set, but the elevation coefficients were significant and
are available on the TEX-CUP website for both starboard and port Sensorium
antennas at both L1 and L2.  Application of these coefficients reduces the
standard deviation of L1 and L2 undifferenced carrier phase residuals by 11\%
and 15\%, respectively, in the 2-day PCV calibration data set.  Coefficients
for the L5 frequency will be posted to the TEX-CUP website in the future.

\subsubsection{GNSS Code Phase Biases}
Differential code phase biases arise in GNSS receivers due to dissimilar
frequency paths and dissimilar autocorrelation functions
\cite{montenbruck2014differential}.  Thus, a bias may arise between GPS L1 C/A
and GPS L2C code phase measurements even though the signals have similar
autocorrelation properties, and between GPS L1 C/A and Galileo E1 measurements
even though the signals have identical center frequencies.  A similar bias
exists at each GNSS satellite.

Monthly estimates of the satellite-side biases are available from the Center
for Orbit Determination in Europe (CODE)\footnote{See
  \url{http://ftp.aiub.unibe.ch/CODE/}}.  Once these are applied, it is
straightforward to estimate the receiver-side biases relative to a reference
signal, usually taken to be GPS L1 C/A.  During the 10-minute stationary
periods that bookend each TEX-CUP data interval, a GPS L1 C/A-only CDGNSS
solution can be obtained for each of the Sensorium's antennas.  Due to the
short Sensorium-to-reference baseline (less than 1 km during these stationary
segments), and to averaging over the 10-minute period, this solution is
accurate to better than 1 cm.  Once obtained, this solution can be used as a
truth constraint on the antennas' location.  Next, a high-accuracy ionospheric
model such as the final TEC grid of the International GNSS Service
\cite{hp_igs_vtec_2009,narula2018accurate} is applied to compensate for
ionospheric delays in code phase.  Finally, the receiver's differential code
phase biases are estimated by averaging pseudorange residuals for each signal
when the antennas are constrained to their known location.

\subsection{Extrinsic Calibration}
\label{sec:extrinsic-calibration}

\begin{figure*}[htb!]
  \centering
  \includegraphics[width=0.8\linewidth] {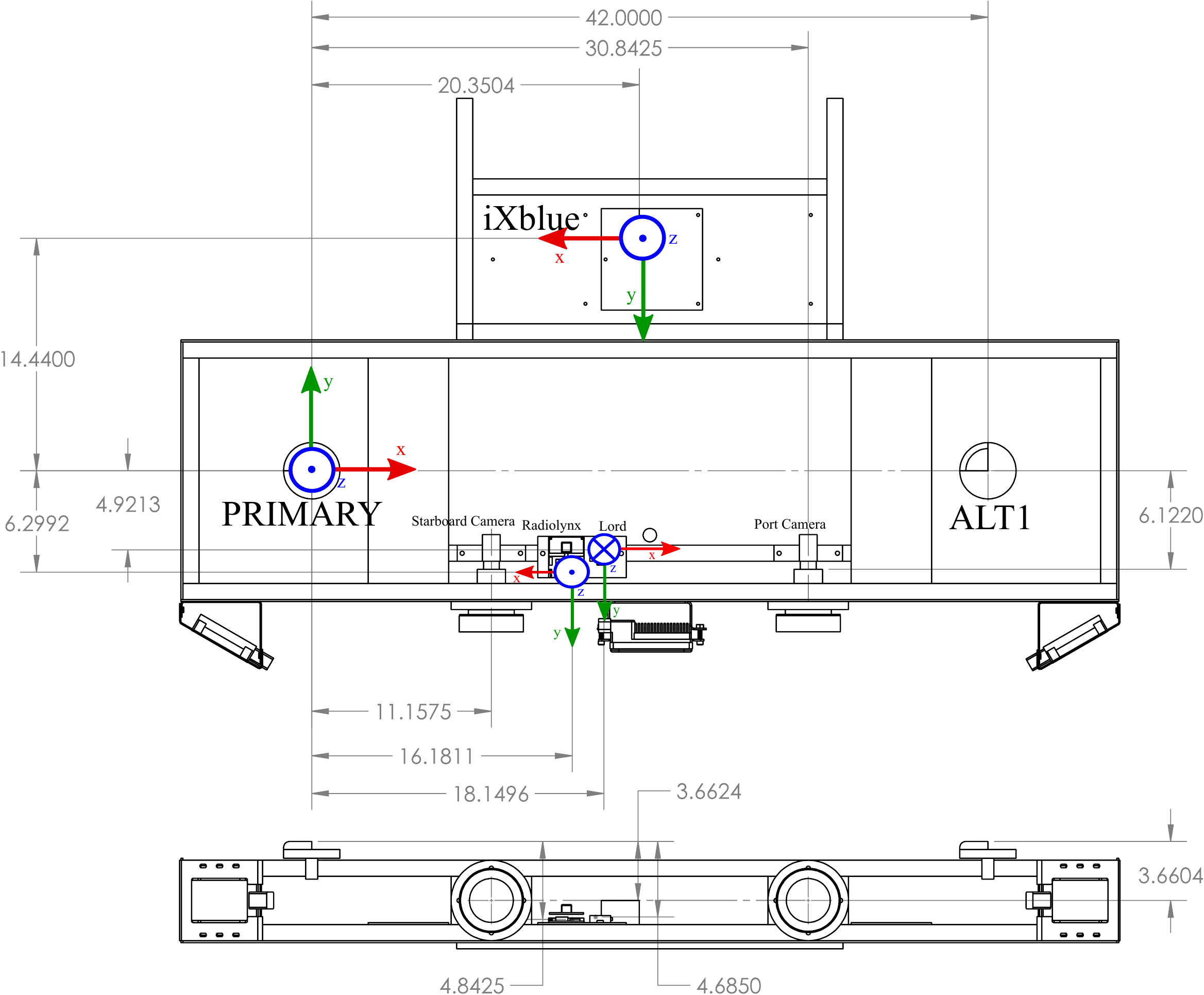}
  \caption{Computer drawing showing the position and orientation of the sensors
  used in this dataset from a top view (top panel) and a front view (bottom
  panel). Measurements are provided in inches. Note that the
  indicated coordinate axes are assumed to be perfect \ang{90} rotations from
  each other.  The $\odot$ symbol denotes an out-of-page axis, while the
  $\otimes$ symbol denotes an in-to-page axis.}
  \label{fig:extrinsic-calibration}
\end{figure*}

Extrinsic calibration involves estimation of the relative positions and
orientations of different sensors involved in sensor fusion.
Fig.~\ref{fig:extrinsic-calibration} shows the coordinate frames involved in
the dataset, and lever arm measurements between these frames. The sensor mounts
are machined to maintain \ang{90} rotations between the sensors. However, it
is possible that the tolerances involved in this process may not be sufficiently
accurate for high-precision positioning.  Accordingly, the provided extrinsic
parameters should be considered as initial estimates to an online calibration
procedure.

Note that the camera calibration data captures mentioned above may be used to
estimate the extrinsic parameters between the two cameras. As noted before,
these parameters have been observed to vary due to platform vibrations and must
ideally be tracked in real time.

\subsection{Synchronization}
\label{sec:sync}

Sensorium IMU and camera measurements are synchronized to GPS time. The Bosch
IMU is built-in to the RadioLynx board, enabling direct synchronization to the
RadioLynx sample clock, and by extension to GPS time. The LORD MicroStrain IMU
accepts a PPS signal from the u-blox receiver and GPS week and whole seconds
over USB from the software-defined GNSS receiver running on the Sensorium
computer. The synchronization to GPS time is handled internally by the LORD
IMU. Similarly, the ATLANS-C IMU measurements and fused ground truth trajectory
are internally synchronized to GPS time (but reported in UTC time).

The Sensorium computer clock is itself synchronized to GPS time to within less
than a millisecond by pointing the computer's NTP client to the GPS time
reported by the software receiver running on the machine. This enables the
Sensorium computer to timestamp any sensor data with sub-millisecond accuracy
to GPS time.

The Basler cameras in the Sensorium accept an external hardware trigger to
capture images. The trigger is generated by the RadioLynx board at
$\approx$\SI{10}{\hertz} in synchronization with the sampling clock ticks. As a
result, the trigger provided to the cameras can be traced back to the GNSS
sample recorded by the RadioLynx front end. There are two major  sources of
delay that may be taken in to account when processing camera images. First,
after receiving the hardware trigger, the cameras expose the sensor for a
variable amount of time, depending on the lighting conditions. Fortunately, the
Basler API provides access to the exposure time for each image. The provided
dataset annotates each individual image with the exposure time reported by the
camera.  Second, the images are timestamped by the Sensorium computer when
these images are received over the local Ethernet connection. The data transfer
time from the camera to the computer is typically very stable since no other
devices are on the network, and may be estimated as a constant parameter in
real time, if necessary.

\section{Summary \& Future Extensions}

A GNSS-based precise positioning benchmark dataset collected in the dense urban
center of Austin, TX has been introduced. With provision of raw wideband IF
GNSS data along with tightly synchronized raw measurements from multiple IMUs
and a stereoscopic camera unit, the authors hope that the precise GNSS
positioning community will benefit from testing their techniques on a
challenging public dataset. In the near future, the authors hope to offer a
benchmarking service similar to the KITTI benchmark
suite~\cite{Geiger2013IJRR}, providing the opportunity for researchers to
publicly compare precise urban positioning methods. The dataset will soon be
extended to include wideband GNSS IF data collected in several other urban centers
around the world.

\bibliographystyle{IEEEtran} 
\bibliography{pangea}

\balance

\end{document}